\documentclass[conference]{IEEEtran}
\usepackage{ifpdf}

\usepackage{cite}

\ifCLASSINFOpdf
\else
\fi
\usepackage{amsmath}
\usepackage{array}
\usepackage{fixltx2e}

\usepackage{stfloats}
\usepackage{url}

\usepackage{booktabs}
\usepackage{multirow}
\usepackage{multirow}
\usepackage{longtable}
\usepackage{graphicx}
\usepackage{bm}
\usepackage{algorithm}
\usepackage[]{algpseudocode}
\usepackage[switch]{lineno}
\usepackage{caption}
\usepackage{color}

\begin{document}
%
\title{Interpretable Real-Time Win Prediction for Honor of Kings -- a Popular Mobile MOBA Esport}

\author{\IEEEauthorblockN{Zelong Yang}
\IEEEauthorblockA{Tsinghua-Berkeley Shenzhen Institute\\Tsinghua University\\
Email: yangzelong14@gmail.com}
\and
\IEEEauthorblockN{Zhufeng Pan}
\IEEEauthorblockA{University of California, Los Angeles\\
Email: panzhufeng@cs.ucla.edu}
\and
\IEEEauthorblockN{Yan Wang}
\IEEEauthorblockA{AI Lab\\ Tencent Inc.\\
Email: brandenwang@tencent.com}
\and
\IEEEauthorblockN{Deng Cai}
\IEEEauthorblockA{The Chinese University of Hong Kong\\
Email: thisisjcykcd@gmail.com}
\and
\IEEEauthorblockN{Xiaojiang Liu}
\IEEEauthorblockA{AI Lab\\ Tencent Inc.\\
Email: kieranliu@tencent.com}
\and
\IEEEauthorblockN{Shuming Shi}
\IEEEauthorblockA{AI Lab\\ Tencent Inc.\\
Email: shumingshi@tencent.com}
\and
\IEEEauthorblockN{Shao-Lun Huang}
\IEEEauthorblockA{Tsinghua-Berkeley Shenzhen Institute\\Tsinghua University\\
Email: twn2gold@gmail.com}}


%


\maketitle

\begin{abstract}
With the rapid prevalence and explosive development of MOBA esports (Multiplayer Online Battle Arena electronic sports), much research effort has been devoted to automatically predicting game results (win predictions). 
\textcolor{black}{While this task has great potential in various applications, such as esports live streaming and game commentator AI systems, 
previous studies fail to investigate the methods to \textit{interpret} these win predictions.} 
\textcolor{black}{To mitigate 
this issue, we collected a large-scale dataset that contains real-time game records with rich input features of the popular MOBA game \emph{Honor of Kings}.} For interpretable predictions, we proposed a \textit{Two-Stage Spatial-Temporal Network} (\textit{TSSTN}) that can not only provide accurate real-time win predictions but also attribute the ultimate prediction results to the contributions of different features for interpretability. Experiment results and applications in real-world live streaming scenarios showed that the proposed TSSTN model is effective both in prediction accuracy and interpretability.


\end{abstract}


%
\IEEEpeerreviewmaketitle

\section{Introduction}

MOBA esports have become increasingly popular. For example, one of the world's three most popular and highest-grossing MOBA games, \emph{Honor of Kings} (\emph{HoK}, whose international version is named \emph{Arena of Valor}), is attracting more than 80 million daily active players and 200 million monthly active players\footnote{\url{https://en.wikipedia.org/wiki/Honor_of_Kings}}. Other popular MOBA games such as \emph{DotA II} and \emph{LoL} also have tens of millions of players worldwide. In the last \emph{King Pro League} Fall 2019 tournament for \emph{HoK}, the prize pool reached 1.13 million dollars\footnote{\url{https://liquipedia.net/arenaofvalor/King_Pro_League/2019/Fall}}. Due to the enormous popularity of MOBA esports, related industries such as esports live streaming and game commentator AI systems have emerged and increasingly generate huge amounts of profit. 
\textcolor{black}{Under these circumstances, much research 
\cite{kinkade2015dota,conley2013does,kalyanaraman2014win,semenov-et-al:performance,yang2016real,wang2018predictive,hodge2017win,makarov-et-al:predicting,hodge2019win,wang2018outcome,song2015predicting} has been done on automatic MOBA esports win predictions.}
\textcolor{black}{Among these works, \cite{kinkade2015dota,conley2013does,kalyanaraman2014win,semenov-et-al:performance,song2015predicting} primarily focus on pre-game win predictions using pre-game features such as hero selections and roles of heroes. Other works \cite{yang2016real,wang2018predictive,hodge2017win,makarov-et-al:predicting,hodge2019win,wang2018outcome} introduce real-time (during match) features such as team-kills and heroes' experience to predict the outcomes of MOBA games. For example, Yang et al.~\cite{yang2016real} use three real-time features (gold, experience, and death) to predict the winner of \emph{DotA II}.}

\begin{figure}[t]
\centering
\includegraphics[scale=0.26]{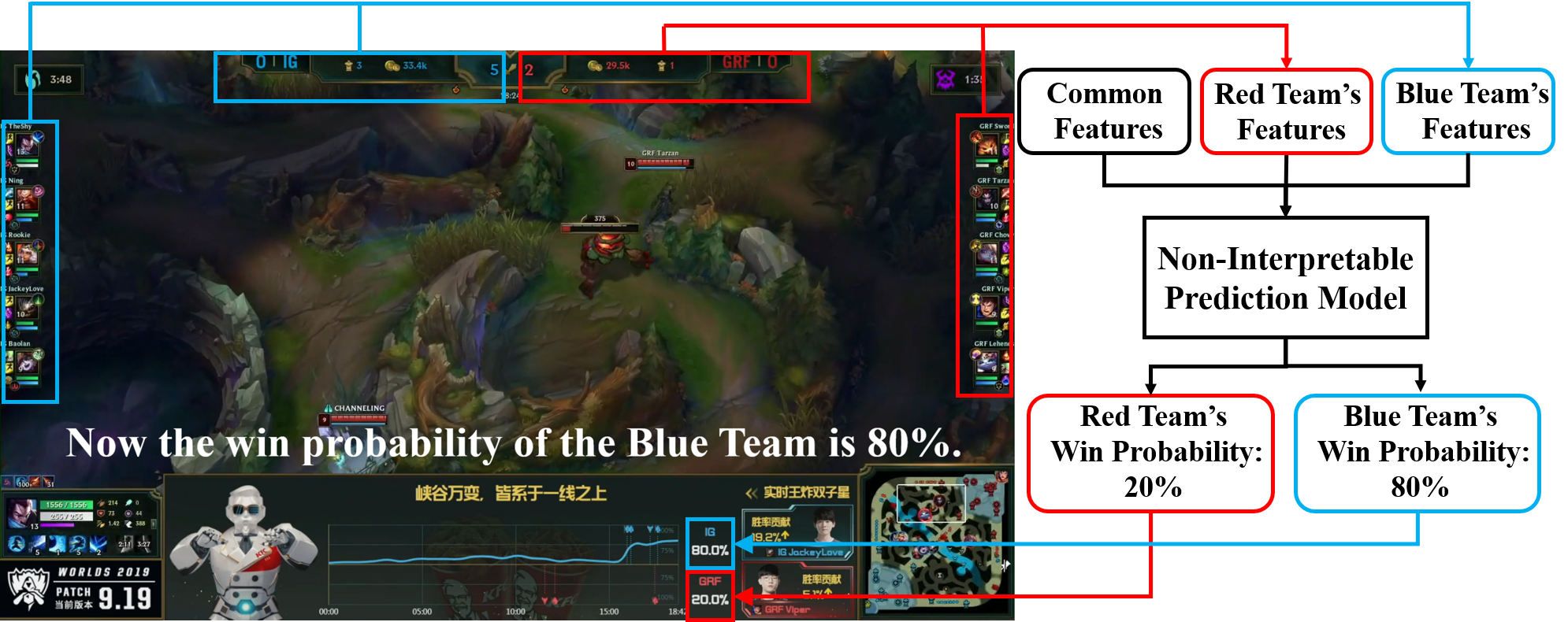} 
\caption{Application of a non-interpretable prediction model in the \emph{LoL} 2019 World Championship live streaming.  Due to the lack of interpretability, the commentators can merely make simple comments like ``\emph{Now the win probability of the Blue Team is 80\%}''.}
\label{fig:kpl}
\end{figure}



\textcolor{black}{Although some progress has been made, existing studies on MOBA esports win predictions suffer from a major limitation that they fail to investigate the possibility to interpret these win predictions.} The lack of interpretability dramatically limits their applications in industry scenarios such as esports live streaming. As one can see in Figure \ref{fig:kpl}, the non-interpretable prediction model used in the \emph{LoL} 2019 World Championship can only give the win probabilities of the two teams (80\% and 20\%). \textcolor{black}{Since the game commentators are not aware of the underlying reasons for the prediction results, they can merely make some simple comments such as ``\textit{Now the win probability of the Blue Team is 80.0\%}'' based on the win probabilities, which is rather dull and non-informative.} \textcolor{black}{In contrast, once we can provide more interpretable predictions such as: \textit{\{WinProbBlue: 80.0\%, Factor 1: towers, Factor 2: team compositions\}}, the commentators can be aware that it is the difference in tower-counts and team compositions that results in these win predictions. With this information, more insightful comments can be made such as: ``\textit{The Blue Team has destroyed three towers in the past five minutes and thus accumulated a huge advantage in terms of vision. Besides, with a more powerful team composition, they now have a very high win probability of 80.0\%.}''}

To address the above problems, we collected a large-scale dataset that contains abundant real-time game records of the popular MOBA game \emph{HoK}. A critical characteristic of these game records is that they are played by highly-skilled human players (top 1\% according to the game leaderboard). Since matches with higher skill levels result in fewer random factors~\cite{yang2016real,conley2013does}, the collected dataset is more predictable and thus can result in more reliable models. The dataset covers 184,362 real human games, and the real-time game records are collected every 30 seconds, generating 5,253,661 data-frames in total. For each data-frame, the most discriminative and human-perceptible features are included, such as teams' accumulated gold, kill-counts, and tower-counts.

With the above collected dataset, we propose a \textit{Two-Stage Spatial-Temporal Network} (\textit{TSSTN}), which gives not only the accurate win predictions but also human-interpretable intermediate results. \textcolor{black}{In the first \textit{Spatial-stage}, features are grouped into six distinct \textit{feature groups} (\textit{Gold}, \textit{Kill}, \textit{Tower}, \textit{Wild Resource}, \textit{Soldier}, and \textit{Heroes}) and projected onto six separate representation spaces.} \textcolor{black}{These are the six most important features suggested by senior professional commentators.} Upon these spaces, we build six \textit{Spatial-models} that take individual feature groups as input only and make the respective win predictions. 
\textcolor{black}{Then, 
after a deep investigation on \textit{HoK} and other MOBA games, we find that the same features}
\textcolor{black}{are not invariantly important throughout the games. For example, in the beginning of the games, the team compositions are the most vital features to the win predictions; however, in the middle of the games, gold amount may become the most important feature in terms of win predictions.} To better model this ``temporal'' characteristic, in the second stage (\textit{Temporal-stage}) we assign six time-variant weights to the six Spatial-models, and the ultimate win prediction is made through the weighted combination of Spatial-models’ outputs. 

The proposed TSSTN model is interpretable in that it separates the effects of the values and importance of features to decouple their contributions: 1) The Spatial-models’ outputs are independent win predictions based on single feature groups, which indicate the two teams' \textit{win-or-lose likelihood} in the domains of different feature groups (Spatial-domains); 2) The six time-variant weights of the Temporal-stage represent the \textit{relative importance} of different feature groups at specific time-points; and 3) The product of the Spatial-model’s output and its time-variant weight denote the \textit{contribution} of the single feature group to the ultimate win prediction. 

\begin{figure}[t]
\centering
\includegraphics[scale=0.38]{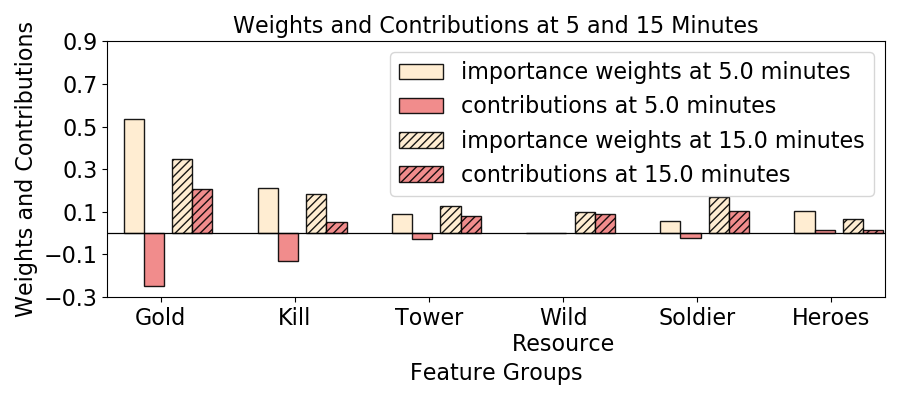}
\caption{Intermediate prediction results in an \emph{HoK} game. This figure shows the importance weights and contributions of six feature groups in the example game at 5 minutes and 15 minutes.}
\label{fig:figure1}
\end{figure}

It is worth noting that high importance does not necessarily lead to a large contribution. \textcolor{black}{For example, team gold is a high-importance feature at most time-points, but its contribution will still be insignificant if the two teams have approximately similar amounts of gold at the current time-point.} On the other hand, features with relatively low importance could still make a non-negligible contribution if the Spatial-model's distinct prediction output is high enough. \textcolor{black}{For instance, in the game shown in Figure \ref{fig:figure1}, the importance weights at 15.0 minutes is the fifth out of six (10.1\%), while its contribution to the current game ranks the third (16.6\%).}

Figure \ref{fig:figure1} illustrates the interpretable intermediate results given by the TSSTN model. As shown, our system gives information about the importance weights and contributions of the six feature groups to explain the underlying reasons for the ultimate predictions. At 5.0 minutes, the most contributed feature groups are \textit{Gold}, \textit{Kill}, and \textit{Tower}, as the pure coral bars show; at 15.0 minutes, the most contributed features change to \textit{Gold}, \textit{Soldier}, and \textit{Wild Resource}, as the striped coral bars show. In this example, the \textit{Gold} feature's importance and contribution decrease as the match progresses, while other features' (such as \textit{Tower} and \textit{Soldier}) contributions and importance increase as the match proceeds.

Experiments show that the TSSTN model gives accurate real-time win predictions as well as human-interpretable intermediate results for the popular MOBA game \emph{HoK}. Besides, the TSSTN model has already been implemented in a real-world AI commentator system to provide human-interpretable prediction results.

The contributions of our work are summarized as follows:
\begin{itemize}
    \item A large-scale \emph{HoK} dataset for real-time MOBA esports win prediction is collected and utilized to train reliable prediction models.
    
    \item A \textit{Two-Stage Spatial-Temporal Network} (\textit{TSSTN}) is proposed to make real-time interpretable win predictions by attributing the predictions to the contributions of different features.
    
    \item  Experiment results and applications in a real-world AI commentator system demonstrate the effectiveness and interpretability of the proposed TSSTN model.
\end{itemize}

\section{Related Works}
Due to the high demand for game result forecasts, much research has been conducted on predicting the results of games, both for traditional sports and MOBA esports. In this section, we introduce some related work on the topic of game result predictions, including traditional sports win predictions, MOBA esports pre-game win predictions, and MOBA esports real-time win predictions.

\subsection{Traditional Sports Win Prediction}
Due to the high demand from related industries, many studies have aimed to predict the winners of traditional sports, such as basketball~\cite{kayhan-watkins:data,kayhan2019predicting,song2020making}, football~\cite{delen-et-al:comparative,leung2014sports}, and others~\cite{bhawkarpredicting,valero2016predicting}. In these works, the authors first collect their datasets containing pre-game or real-time information, then build models to predict the results based on their selected features. These models have proven to be useful to some extent, but two issues need to be stressed. Firstly, to make sure that the games are of the same skill level, the datasets should contain games in the same professional competitions, such as the \emph{NBA} (\emph{National Basketball Association}) for basketball and the \emph{NFL} (\emph{National Football League}) for football. However, the amount of such sports competitions' data is far too insufficient to train reliable models, considering the limited number of matches every year. Secondly, there are too many off-game variables in traditional sports, such as weather conditions.\footnote{\textcolor{black}{Esports also have many off-game variables that can affect the results of the games, such as the player being tired, ill, under stress, and mood. However, we claim that these factors can be well reflected by the in-game features such as the gold-amount and the number of towers. In this sense, the win predictions for esports are more generalizable than traditional sports predictions.}} Therefore, it is hard to comprehensively and objectively choose the suitable features.

\subsection{MOBA Esports Pre-Game Win Prediction}
MOBA esports games, on the other hand, are naturally suitable for the task of result prediction. For one thing, huge amounts of structured 
game data can be easily collected from the game database to train reliable models. Also, the variables in MOBA esports are relatively few and easy to obtain. With the rapid development and prevalence of MOBA esports, much research has been conducted on predicting the results of MOBA esports games, such as~\cite{makarov-et-al:predicting,hodge2019win}. Among these works, \cite{conley2013does} first started research on MOBA esports result predictions, then \cite{kalyanaraman2014win,song2015predicting,kinkade2015dota} followed this work to make pre-game win predictions. Semenov et al.~\cite{semenov-et-al:performance} and Wang et al.~\cite{wang2018outcome} suggested several machine learning methods to predict MOBA esports using pre-game information such as team compositions. Chen et al.~\cite{chen-et-al:modeling} studied the synergy and restrain relations among heroes. These works, however, lack real-time data, a factor limiting their usages.

\subsection{MOBA Esports Real-Time Win Prediction}

Yang et al.~\cite{yang2016real} first used real-time data to predict the results of MOBA esports. However, their training data and selected features were insufficient (just less than one-third of the training data contained real-time information, and only three real-time features were collected). Besides, the results of this work were non-interpretable, further limiting their use. \textcolor{black}{Following this work, many works \cite{wang2018predictive,hodge2017win,makarov-et-al:predicting,hodge2019win,wang2018outcome} used richer real-time features to give win predictions for \emph{DotA II} and \emph{LoL}.} Hodge et al.~\cite{hodge2017win} used the game data of different skill levels to mitigate the inefficiency of training data. \cite{katona2019time} aimed to predict events inside the games such as hero deaths. Demediuk et al.~\cite{demediuk-et-al:role} used unsupervised learning technologies to explore the roles of heroes in MOBA esports.

Another type of research on esports win predictions is to train game AI using reinforcement learning technologies, such as~\cite{sun-et-al:tstarbots} for \emph{StarCraft II}, \cite{ye-el-al:mastering,zhang-et-al:hierarchical} for \emph{Honor of Kings}, and~\cite{OpenAI:OpenAI_dota} for \emph{DotA II}. Trained appropriately, value-based reinforcement learning models are capable of giving real-valued win prediction results. Although these game AI models can give real-time win predictions, they are black-box models lacking in interpretability.

Based on these previous works, our work utilizes sufficient MOBA esports data with diverse real-time features and designs a TSSTN model to give reliable and interpretable prediction results. Our work has already been implemented in a real AI commentator system for \emph{HoK}. 
\textcolor{black}{To the best of our knowledge, 
our work is a pioneering attempt to realize interpretable prediction models for MOBA esports win predictions.}

\begin{table}[t]
\caption{An example data-frame at time 9.5 minutes.}
\centering
\resizebox{80mm}{!}{
\begin{tabular}{@{}llllllllllll@{}}
\toprule
 \multicolumn{10}{l}{time: 9.5} &  \\ \midrule
 \multicolumn{5}{l|}{win side: $\cdots$} & \multicolumn{5}{l}{middleTowerCnt: 3}  \\
 \multicolumn{5}{l|}{lose side:} & \multicolumn{5}{l}{crystalTowerCnt: 3}  \\
  & \multicolumn{4}{l|}{hero id:} & \multicolumn{5}{l}{AD:}  \\
  &  & \multicolumn{3}{l|}{52, 96, 56, 23, 31} &  & \multicolumn{4}{l}{alive: true}  \\
  & \multicolumn{4}{l|}{hero:} &  & \multicolumn{4}{l}{id: 31}  \\
  &  & \multicolumn{3}{l|}{52:} & \multicolumn{5}{l}{Raider: $\cdots$}  \\
  &  &  & \multicolumn{2}{l|}{profession: mage} & \multicolumn{5}{l}{AP: $\cdots$}  \\
  &  &  & \multicolumn{2}{l|}{deadCnt: 3} & \multicolumn{5}{l}{overlord: 0}  \\
  &  &  & \multicolumn{2}{l|}{assistCnt: 1} & \multicolumn{5}{l}{darkTyrant: 0}  \\
  &  &  & \multicolumn{2}{l|}{killCnt: 0} & \multicolumn{5}{l}{numOverlord: 0}  \\
  &  &  & \multicolumn{2}{l|}{gold: 3338} & \multicolumn{5}{l}{numDarkTyrant: 0}  \\
  &  & \multicolumn{3}{l|}{96: $\cdots$} & \multicolumn{5}{l}{numTyrant: 0}  \\
  &  & \multicolumn{3}{l|}{56: $\cdots$} & \multicolumn{5}{l}{numRedBuf: 0}  \\
  &  & \multicolumn{3}{l|}{23: $\cdots$} & \multicolumn{5}{l}{numBlueBuf: 0}  \\
  &  & \multicolumn{3}{l|}{31: $\cdots$} & \multicolumn{5}{l}{soldierDist:}  \\
  & \multicolumn{4}{l|}{teamGold: 23884} &  & \multicolumn{4}{l}{lower: 8544}  \\
  & \multicolumn{4}{l|}{kill: 6} &  & \multicolumn{4}{l}{upper: 8932}  \\
  & \multicolumn{4}{l|}{participation: 0.67} &  & \multicolumn{4}{l}{middle: 7273}  \\
  & \multicolumn{4}{l|}{towerCnt: 7} &  & \multicolumn{4}{l}{ }  \\ \bottomrule
\end{tabular}
}
\label{tab:data}
\end{table}

\section{Dataset}
\label{sec:Dataset}

\begin{figure*}[t]
\centering
\includegraphics[scale=0.47]{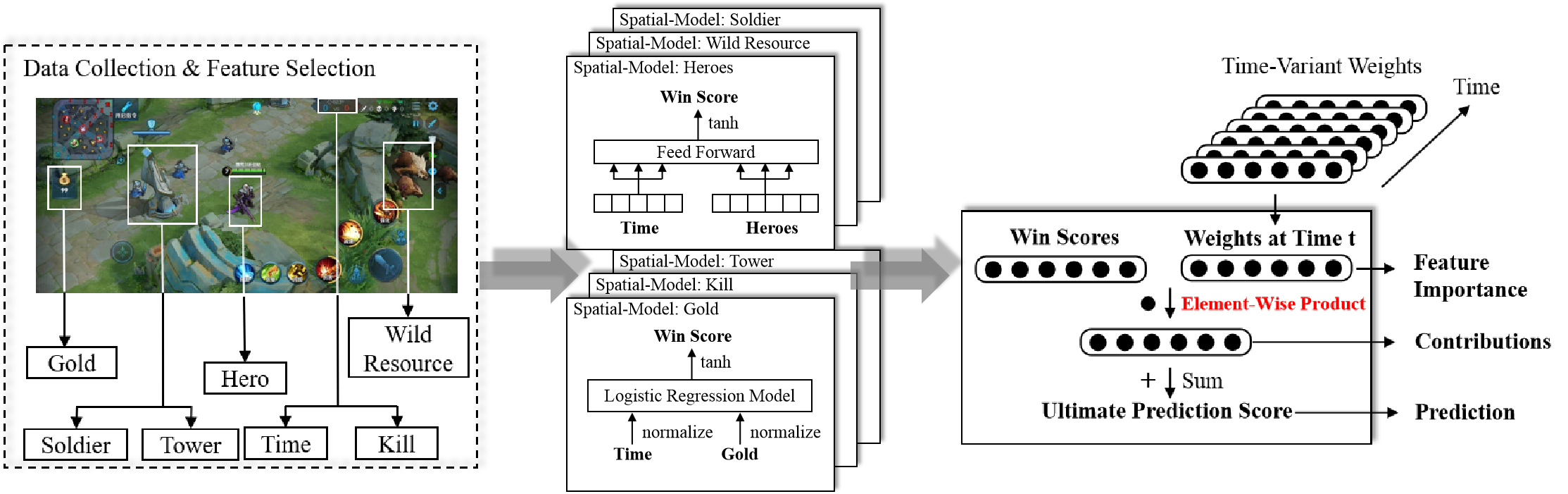}
\caption{The flowchart of the proposed TSSTN model. The left part displays the data collection and feature selection procedures. The middle part is the Spatial-stage. Each Spatial-model is designed for one feature group to give individual prediction (win-score). The right part shows the Temporal-stage. The time-dependent importance weights combine with Spatial-models' outputs to give the ultimate win prediction. The middle part (Spatial-stage) consists of two kinds of Spatial-models: feed-forward networks and logistic regression models. For feed-forward Spatial-models, the features (including time) are converted into embedding vectors as input; for logistic regression Spatial-models, the input features are scalars. \textcolor{black}{In the figure, the different numbers of boxes for ``Time'' and ``Heroes'' represent the different length of these vectors.}}
\label{fig:flowchart}
\end{figure*}

\textcolor{black}{\emph{HoK} (along with its international version \emph{Arena of Valor}) is a mobile MOBA game that enjoys great popularity globally.} In \emph{HoK}, two teams of five heroes combat against each other to destroy the ``\textit{Crystal}'' (home base) of the rival team for victory. \textcolor{black}{A detailed game introduction can be found in \cite{wikipedia:HonorofKings}.}

\textcolor{black}{Cooperating with the publisher of the \emph{Honor of Kings} game, we collect the anonymized information\footnote{\textcolor{black}{Data without the privacy information such as players' names, account IDs, and game history.}} of all the 184,362 games that happened on a randomly chosen day in June 2019 from the game-core data of the official game database. Game-core data is the original data form with which we can reconstruct the replays of the games and extract the necessary game features.} Prior experiments for data scale indicate that one day's data is sufficient for our task. Please note that to ensure the quality of games, only ``\textit{Conquerors}''-level games - whose participants are top 1\% players - are collected (according to~\cite{yang2016real,conley2013does}). To reduce the redundancy of the dataset, we extracted the features every half minute. This means that a 15-minute game will transform to 30 data-frames in the collection process, and we obtained 5,253,661 data-frames in total. In each data-frame, 96 features from the game process were extracted, including sufficient information that we think is helpful for the win prediction.\footnote{\textcolor{black}{As Table \ref{tab:data} shows, the 96 features include 48 features for the win and the lose side: heroes' IDs (1); profession, dead count, assist count, kill count, and gold of the five heroes (5 $\times$ 5); team gold (1); team's kill count (1); participation (1); tower count (1); Middle Tower count (1); Crystal Tower count (1); alive status and IDs for AD, Raider, and AP (2 $\times$ 3); Overlord buff status (1); Dark Tyrant buff status (1); number of killed Overlord (1); number of killed Dark Tyrant (1); number of killed Tyrant (1); number of killed Red and Blue Buff (1 $\times$ 2); soldiers' distances to the crystal (1 $\times$ 3).}} Aside from some obviously useful information such as gold, kill-count, and tower-count of each team, we also calculated some potentially helpful features, such as soldiers' distances to the rival's ``Crystal'', buff information, and wild resource occupancy. An illustration of one data-frame is shown in Table \ref{tab:data}, where fields of different layers are separated using indents, and repeated fields are omitted using ellipses.


\section{Model}
\label{sec:Model}

In this section, we introduce the Two-Stage Spatial-Temporal Network (TSSTN) model for interpretable real-time MOBA esports win predictions. In the first \textit{Spatial-stage}, we categorized the features into six feature groups: \emph{Gold}, \emph{Kill}, \emph{Tower}, \emph{Wild Resource}, \emph{Soldier}, and \emph{Heroes}. Then we projected them onto different representation spaces. Upon each space, a \textit{Spatial-model} was built to compute a \textit{win-score} $s_i \in[-1.0,~1.0]$ which represents the win-or-lose likelihood prediction based on this feature group. Since each Spatial-model was unaware of the features in other feature groups, its output (win-score) could be regarded as the independent win prediction of the team in each specific feature domain. Moreover, these spaces (feature groups) were neither equally important with each other (in spatial-dimensions) nor invariantly important throughout the games (in temporal-dimensions). For example, feature group \textit{Soldier} (soldiers' distances to the rival Crystal) became increasingly important because soldiers became increasingly potent as the games proceeded. To better model the temporal variance of feature groups' relative importance, in the second \textit{Temporal-stage} we assigned a time-variant and learnable weight to each Spatial-model. Finally, the inner product of the win-score vector $\boldsymbol{s}=[s_1, s_2, ..., s_6]$ and the time-dependent importance weight vector $\boldsymbol{w}^t=[w_1^t, w_2^t, ..., w_6^t]^{T}$ was the ultimate prediction result $F^{t}$. Figure \ref{fig:flowchart} demonstrates the flowchart of the proposed TSSTN model.

In this two-stage architecture, some intermediate results can help interpret the ultimate win predictions. 1) The weight vector $\boldsymbol{w}^{t}$ at time $t$ reveals the relative importance of the feature groups at that time. It varied over time because feature groups were not invariantly important in different game processes. 2) The product $w_i^t\cdot s_i$ of feature group $x_i$'s importance weight $w_i^t$ and win-score $s_i$ directly contributed to the ultimate win prediction $F^{t}$ as $F^{t} =\sum_i w_i^t\cdot s_i$. Therefore, we refer to the value of $w_i^t\cdot s_i$ as the \textit{contribution} of feature group $x_i$. This contribution revealed the effects of the two teams' feature differences and features' importance on the ultimate prediction result. Based on this scheme, we can easily explain how the ultimate win prediction $F^{t}$ is made according to the above interpretable information. \textcolor{black}{In many previous studies, this information was hidden in the entangled parameters of the ``black-box'' models.}

\subsection{Feature Selection}
\label{sec:feature}


\textcolor{black}{To choose the most suitable features from the dataset, we invite senior \emph{HoK} commentators to select the most relevant features for win predictions based on their professional experience. Based on their suggestions, we select 11 types of features in every data-frame and classify these features into six feature groups}\textcolor{black}{. One may select other features when predicting other MOBA games.}

\begin{itemize}

    \item \textbf{Gold}: The difference in the accumulated amounts of gold between the two teams. Gold can be used to purchase more powerful equipment to enhance the attributes (such as \textit{health-point}, \textit{attack}, and \textit{defense}) of the heroes.
    
    \item \textbf{Kill}: The difference in the kill-counts of the two teams. By killing the rival heroes, extra gold and experience can be obtained to purchase equipment and level up, respectively.

    \item \textbf{Tower}: The difference in the numbers of existing defense towers, including three kinds of towers (front, middle, and Crystal towers). Defense towers can hold back the rival heroes and soldiers as well as provide vision.

    \item \textbf{Wild Resource}: The difference of the wild resources, including two types of wild monsters: \textit{Dark Tyrants} and \textit{Overlords}. By killing these monsters, the whole team obtain buffs that can provide extra gold, experience, and attributes.
    
    \item \textbf{Soldier}: The soldiers' distances to the opponent ``Crystal'', including the distances in the three lanes (top, middle, and bottom lanes). Soldiers are useful in terms of destroying towers, which are unavoidable obstacles to get to the rival ``Crystal''. As the game proceeds, the soldiers will become increasingly potent according to the MOBA game settings.
    
    \item \textbf{Heroes}: \textcolor{black}{The heroes' IDs selected by the players.} \textcolor{black}{As shown in Table \ref{tab:data}, these IDs are in the form of numbers in the dataset.} Since there are certain restrain and synergy relations among heroes, this information can be utilized to improve the prediction accuracy before and during the games.
\end{itemize}

As the input of the TSSTN model, different feature groups were pre-processed in different ways according to their Spatial-models. As the middle part of Figure \ref{fig:flowchart} shows, the first three numerical feature groups were normalized using min and max feature values, and the remaining three categorical feature groups were converted into embedding vectors.

\subsection{Spatial-Stage}

As noted, the Spatial-stage consisted of three logistic regression models and three feed-forward neural networks called \textit{Spatial-models}. We used logistic regression models for numerical feature groups (\textit{Gold}, \textit{Kill}, and \textit{Tower}), and used two-hidden-layer feed-forward neural networks for categorical feature groups (\textit{Wild Resource}, \textit{Soldier}, and \textit{Heroes}). For Spatial-model $i$, the input consisted of its corresponding feature group $x_i$ introduced in Section \ref{sec:feature}, and the current game time $t$. \textcolor{black}{By adding the game time as part of input feature in the Spatial-models, the accuracy of the Spatial-models is improved, so is the overall accuracy of the TSSTN model.} The output of Spatial-model $i$ is a win-score $s_i(x_i,t) \in [-1.0,~1.0]$, representing the win-or-lose likelihood prediction based on the single feature group's value difference.

Note that although adopting more sophisticated Spatial-model architectures for different feature groups will possibly improve the overall prediction accuracy further, it was not the primary purpose of this study. We wanted to focus on the interpretability of MOBA esports win predictions, so we chose simple-architecture networks (logistic regression models and feed-forward networks) for the feature groups. In real-world scenarios, one can deliberately design different model-architectures for different feature groups.

\subsection{Temporal-Stage}

The Temporal-stage consisted of six time-dependent weights, reflecting the relative importance of the six feature groups at different game processes. Given the win-score vector $\boldsymbol{s}(\boldsymbol{x},t)$ at time-point $t$ provided by the Spatial-stage (where vector $\boldsymbol{x}$ represents the six feature groups), the Temporal-stage calculated the ultimate win prediction $F^t$ using the linear combination of importance weights $\boldsymbol{w}^t$ and win-scores $\boldsymbol{s}(\boldsymbol{x},t)$:
\begin{equation}
    F^t = \boldsymbol{w}^t \cdot \boldsymbol{s}(\boldsymbol{x},t) = \sum_i^d  w_i^t \textcolor{black}{\cdot} s_i(x_i,t)
    \label{eqa:win_prediction}
\end{equation}
where $d$ denotes the number of feature groups. Since the weights $\boldsymbol{w}^t$ represent the relative importance of feature groups at time-point $t$, we applied the $softmax$ function to normalize the importance weight $w_i^t$ for each feature group: 
\begin{equation}
    w_i^t=\frac{\exp{\theta_i^t}}{\sum_j \exp{\theta_j^t}},
\end{equation}
so that the weight $w_i^t$ satisfy:
\begin{equation}
    \sum_i^d w_i^t=1,~w_i^t>=0.
\end{equation}

\subsection{\textcolor{black}{Interpretability: Feature Groups' Contributions}}

\textcolor{black}{Interpretable results can then be generated by element-wise multiplying the results of the Spatial-Stage and the Temporal-stage. As the right part of Figure \ref{fig:flowchart} shows, by multiplying the win-scores of the spatial-models and their corresponding importance weights, we get the contributions of the feature groups. After sorting these contributions, we can conclude which feature groups are more important to the current win prediction compared with others. What's more, the summation of all the six contributions is the current win prediction, as shown in Equation \ref{eqa:win_prediction}. For example, as shown in the case of Figure \ref{fig:figure1}, at time 15.0 minutes, the win-scores of the six feature groups are $[0.590,~0.292,~0.612,~0.901,~0.611,~0.231]$ (which is not shown in Figure \ref{fig:figure1}), and the importance weights are $[0.348,~0.184,~0.129,~0.101,~0.172,~0.065]$. After element-wise multiplying the win-scores and the importance weights, the contributions of the six feature groups are $[0.205,~0.054,~0.079,~0.091,~0.105,~0.015]$. By sorting these contributions, we can conclude that feature group 1 (Gold) and feature group 5 (Soldier) are the most important. With this information, human commentators or rule-based commenting systems can generate comments such as ``\textit{The win probability of the Blue Team is 77.5\%, which is due to its advantages in gold amount and soldier positions.}''}

\subsection{Training}

To optimize the parameters in the Spatial-stage and the Temporal-stage, we trained the Spatial-models and importance weight vectors using cross-entropy loss and back-propagation technologies. In experiments, we found little difference between training the two stages simultaneously or separately. In the subsequent experiments, we choose to train the two stages simultaneously.

\section{Experiments}
\label{sec:experiments}
\subsection{Experimental Settings}
\label{Experimental settings}

We conducted experiments on the dataset mentioned above. 10,000 games from the dataset were randomly selected as the test set. 90\% of the remaining 174,362 games were set as the training set and the other 10\% as the validation set.

The parameters of the TSSTN model were as follows. In the Spatial-stage, the Spatial-models for numerical feature groups \textit{Gold}, \textit{Kill}, and \textit{Tower} were logistic regression models with $tanh$ as the non-linear function; the Spatial-models for other categorical groups \textit{Wild Resource}, \textit{Soldier}, and \textit{Heroes} were two-hidden-layer feed-forward neural networks with dimensions (256, 16), (256, 16) and (128, 16), respectively.\footnote{\textcolor{black}{These parameters (number of layers, dimensions of the layers, and dropout parameters) are chosen through a brief parameter sweep, in which we found that parameters in a reasonable range resulted in similar accuracy, and we chose the parameters with the best accuracy.}\label{parameter}} Before the input layer, the feature values (including time) were normalized using min and max values (for logistic regressions) or preprocessed into embedding vectors (for feed-forward networks), as the middle part of Figure \ref{fig:flowchart} shows. For feed-forward network Spatial-models, after each hidden-layer, we used a leaky $ReLU$ function\footnote{\textcolor{black}{The formula of the leaky $ReLU$ is:
\begin{equation}
    \begin{aligned}
    f(x) &= x,~x\geq0,\\
    f(x) &= 0.01 x,~x<0.
    \end{aligned}
\end{equation}
}}
as the non-linear function and added a dropout layer with a rate of $0.2$ to mitigate overfitting.\textsuperscript{\ref{parameter}} After the output layers of the feed-forward network Spatial-models, we used $tanh$ functions to rectify the Spatial-models' outputs (win-scores) to be among $[-1.0,~1.0]$ to represent win-or-lose likelihood.

\subsection{Comparing Methods}

\label{Compared methods}
In order to thoroughly evaluate the aforementioned TSSTN model, we compare it with two baselines: 
\begin{itemize}
\item \textbf{Heuristic:} Intuitively, ``gold'' is one of the most decisive features for the ultimate win predictions, so the \textit{Heuristic} model predicts the game based on the difference of the accumulated gold between the two teams. The team with more gold will be predicted as the winner.

\item \textbf{Fully-Connected:} A fully-connected neural network that takes all the six feature groups as the input. Before the input layer, the features were pre-processed in the same manner as the TSSTN model. We designed four hidden layers for this model with dimensions 1024, 4096, 512 and 64, respectively. After each hidden layer, we added a leaky $ReLU$ function as the non-linear layer and added a dropout layer in order to mitigate overfitting. After the output layer, a $tanh$ function was used to give the ultimate win predictions. The settings of the leaky $ReLU$ layers and dropout rates were identical to those of the TSSTN model introduced in Subsection \ref{Experimental settings}.

\item \textbf{\textcolor{black}{Logistic Regression:}} \textcolor{black}{The Logistic Regression takes all the features as its input. The input is in the same format of the Fully-Connected network.}

\item \textbf{\textcolor{black}{LSTM:}} \textcolor{black}{We use bidirectional LSTM with two recurrent layers. Sequential data with sequence $5$ is used as input. For time 0.0, 2.5, 5.0, and 7.5, the first frame of input data is duplicated to satisfy the sequence-length. The probability of dropout is $0.2$. The size of the hidden state is $32$. After the LSTM, we use a $64$-dimension fully-connected layer and a $tanh$ function to compute the win predictions.}

\end{itemize}

The prediction accuracy of the single-feature-group based prediction models (Spatial-models, such as the prediction model depending on the ``\textit{Heroes}'' features only) are elaborated in Subsection \ref{subsec:FurtherAnalysis}. These Spatial-models are separately discussed since they utilize less input information than the three aforementioned models.

\begin{table}[t]
\caption{The accuracy (\%) of the three prediction models at five equidistant time-points.}
\centering
\resizebox{88.5mm}{!}{
\begin{tabular}{@{}llllllll@{}}
\toprule
 Models & 0 min (std) & 5 min (std) & 10 min (std) & 15 min (std) & 20 min+ (std) \\ \midrule

Heuristic & 50.0 & 69.2 & 78.4 & 73.0 & 63.0 \\
Fully-Connected & 54.7~(0.043) &69.3~(0.023) & 79.3~(0.012) & 75.0~(0.023) & 70.0~(0.057)\\
TSSTN & 54.6~(0.036) & 69.2~(0.005) & 78.5~(0.005) & 73.4~(0.009) & 67.8~(0.032)\\
\textcolor{black}{Logistic Regression} & \textcolor{black}{53.9~(0.007)} & \textcolor{black}{63.8~(0.012)} & \textcolor{black}{78.2~(0.010)} & \textcolor{black}{70.5~(0.008)} & \textcolor{black}{56.6~(0.003)}\\
\textcolor{black}{LSTM} & \textcolor{black}{54.3~(0.021)} & \textcolor{black}{69.4~(0.017)} & \textcolor{black}{78.4~(0.010)} & \textcolor{black}{76.3~(0.005)} & \textcolor{black}{66.5~(0.005)}\\
\bottomrule
\end{tabular}
}
\label{tab:evaluation}
\end{table}

\begin{figure}[t]
\centering
\includegraphics[scale=0.38]{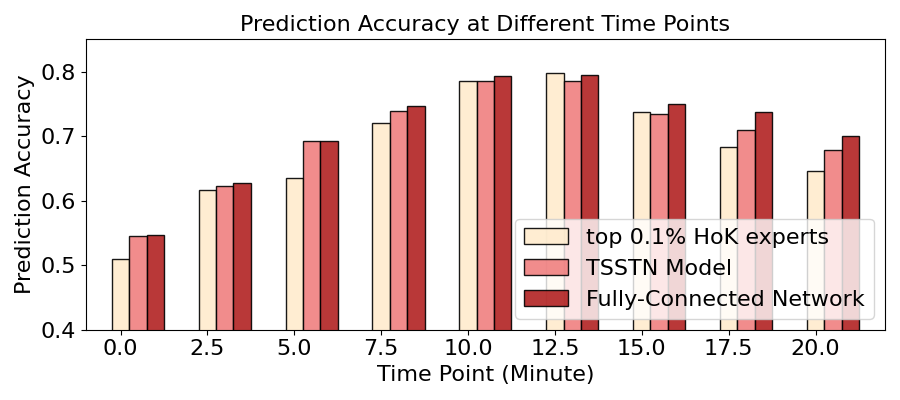}
\caption{The average prediction accuracy of the three \emph{HoK} experts and the two automatic prediction models at nine equidistant time-points.}
\label{fig:annotation}
\end{figure}

\subsection{Evaluation Metrics}
\label{Evaluation metrics}

The prediction accuracy at different game time-points was adopted as the automatic evaluation metrics. Although real-valued winning probabilities can be provided by the TSSTN model and the Fully-Connected Network, we only considered the prediction accuracy of the binary win-or-lose results for the test games. To better evaluate how well the proposed TSSTN model predicts the game-winner, we also recruited three human experts to manually predict the winner of each game given the test game data. All these experts ranked in the top 0.1\% in the ladder tournament of \emph{HoK} according to the game leaderboard. \textcolor{black}{Experts with higher ranks are more likely to predict the games more accurately.} The game experts were given access to the real games so they could explore the whole game states and extract their own features and conclusions, and were asked to predict the winners of the games every 2.5 minutes based on the battle situations and their experience. Note that since it is an impossible task to annotate all the 10,000 games in the test set manually due to the limitation of resources, we randomly chose 200 test games for each time-point for human annotation.

\subsection{Results}

The average prediction accuracy at five equidistant time-points is shown in Table \ref{tab:evaluation}. \textcolor{black}{The Fully-Connected Network and the LSTM achieved the best performance in terms of the prediction accuracy at these five time-points. The prediction accuracy of the TSSTN model was higher than the Heuristic model but a little lower than the Fully-Connected Network and the LSTM, since we sacrificed the coupling information among the six feature groups to some extent for interpretability.} The accuracy of the Heuristic model at 0.0 minute was 50.0\% since when the two teams have equal gold, the Heuristic model predicted the winner stochastically. Please note that human experts' prediction results are not included in Table \ref{tab:evaluation} due to the difference of the test sets (200 games versus 10,000). The comparison between \emph{HoK} experts and the proposed TSSTN model is separately shown in Figure \ref{fig:annotation}.

Figure \ref{fig:annotation} shows the comparisons of the prediction accuracy among human experts and two prediction models, one interpretable model TSSTN and one non-interpretable model Fully-Connected Network. Although the accuracy of the TSSTN model was slightly lower than that of the Fully-Connected Network, it was obviously higher than that of the human experts at time-points 0.0, 17.5, and 20.0. Intuitively, predictions at these three time-points were difficult for human experts because many features that were decisive most of the time were not significant enough at these time-points. For example, ``gold'' is a decisive feature at most time-points. However, at the beginnings of the games (0.0 minute), the two teams had equal gold, and gold was no longer essential after 17.5 minutes because the players had already accumulated enough gold at that time.

\begin{figure*}[t]
\centering
\includegraphics[scale=0.57]{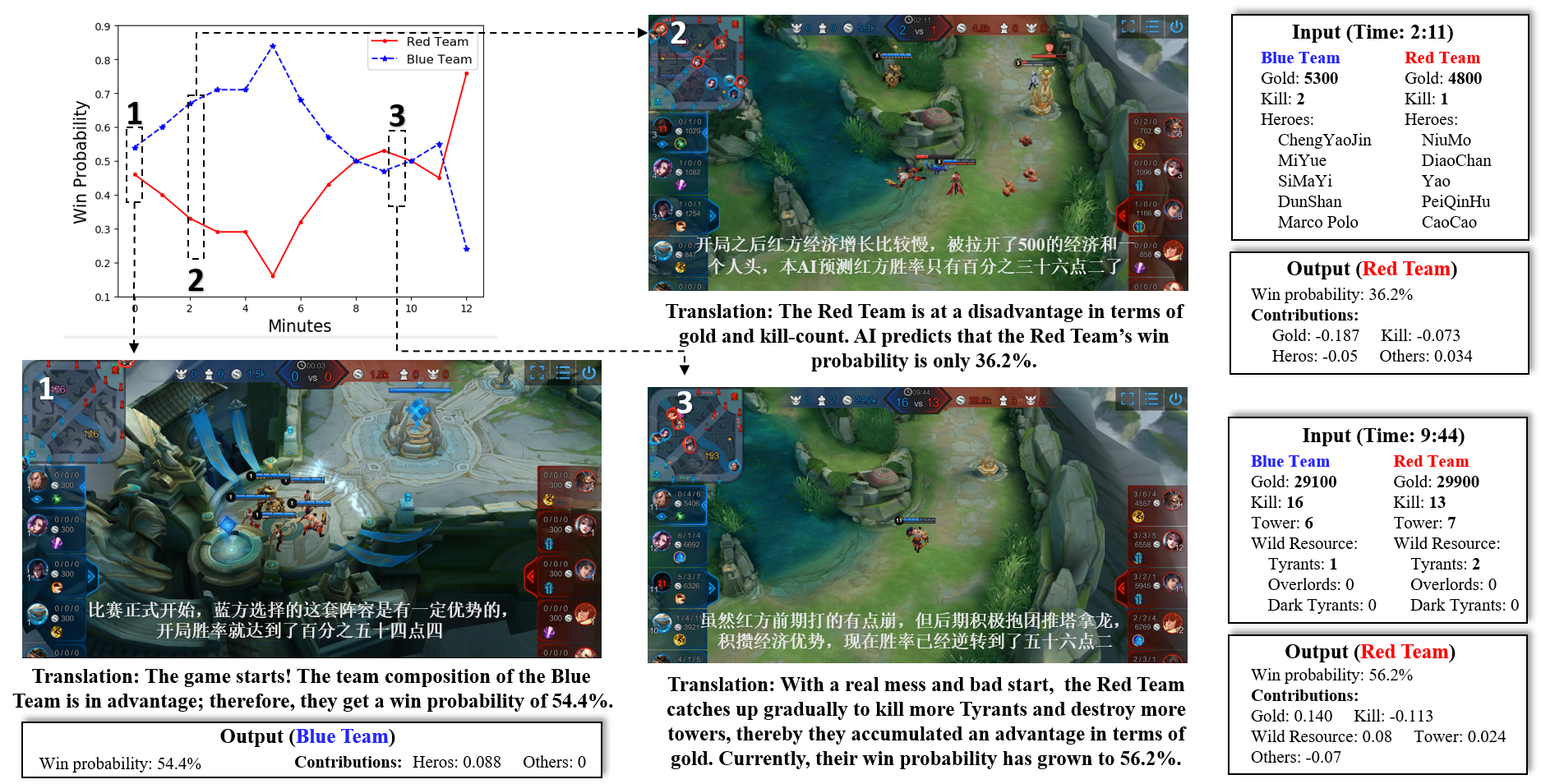}
\caption{An example \emph{HoK} game in which an AI commentator system gives real-time comments as the game progresses. Among these comments, the win prediction related ones are supported by the proposed TSSTN model. The upper left part of this figure shows the win probability curves of the Blue Team and the Red Team throughout the game. The lower left, upper right and lower right parts show three screenshots in which the AI commentator system utilizes the interpretable intermediate results of the proposed TSSTN model to generate its comments.}
\label{fig:example}
\end{figure*}

At time-point 0.0 minute, the \emph{HoK} experts appeared to have low prediction accuracy because before the games started, the human experts could only ``guess'' the results based on prior-game information like team compositions, which was quite inaccurate compared with the prediction results of the automatic methods.

As shown in Table \ref{tab:evaluation} and Figure \ref{fig:annotation}, the prediction accuracy of both the human experts and the automatic methods fell after 12.5 minutes because: 1) In the late-game stages, both the levels and equipment of the two teams reached the maximum; therefore, the results of the games were increasingly affected by random factors like players' accidental mistakes. This fact made the late-stage games harder to predict; and 2) since the games were not equal-duration, games that were easier to predict often ended before late-game stages; therefore the test data-frames were harder to predict at late-game stages time-points, resulting in the noted drop in prediction accuracy.

\subsection{Further Analysis}
\label{subsec:FurtherAnalysis}

\begin{figure}[t]
\centering
\includegraphics[scale=0.38]{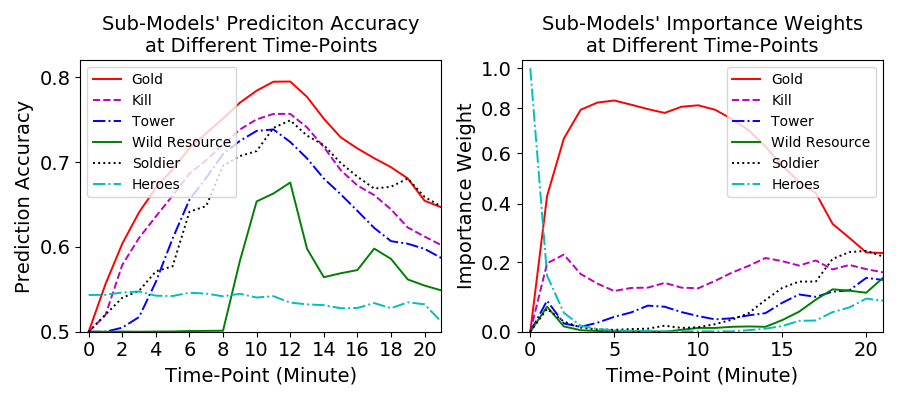}
\caption{Six Spatial-models' average prediction accuracy and importance weights in the first 20 minutes of the games.}
\label{fig:sub-model's weights and accuracy}
\end{figure}

More information about the proposed TSSTN model can be revealed by the six Spatial-models' individual prediction accuracy and the corresponding importance weights, as shown in Figure \ref{fig:sub-model's weights and accuracy}. At time-point 0.0 minute of the games, the only Spatial-model with a non-trivial (larger than 50\%) prediction accuracy was Spatial-model \emph{Heroes}; accordingly, the only non-zero importance weight was also \emph{Heroes}, as shown in the right part of Figure \ref{fig:sub-model's weights and accuracy}. As the games proceeded, the most accurate Spatial-model was \emph{Gold}, which was also reflected in its high importance weights. After 12.0 minutes in the games, both the prediction accuracy and importance weight of \emph{Gold} began to drop. This phenomenon may have occurred because extra gold above some certain thresholds became useless due to the limited package space for equipment in the late-game stages. After 20.0 minutes in the games, the prediction accuracy of the Spatial-model \emph{Soldier} surpassed that of the Spatial-model \emph{Gold}, as was its corresponding importance weight.

The six Spatial-models' importance weights and prediction accuracy changes were consistent with our intuition. For example, since the team composition information (\emph{Heroes}) was fixed throughout each game, the prediction accuracy of the Spatial-model \emph{Heroes} was almost impervious to time, as revealed in the left part of Figure \ref{fig:sub-model's weights and accuracy}. Moreover, since the difference in team compositions had a relatively limited influence on the game results, the corresponding prediction accuracy of Spatial-model \emph{Heroes} was only slightly above 50.0\%. The Spatial-model \emph{Wild Resource} only became valid after 8.0 minutes in the games, since its features ``Overlord'' and ``Dark Tyrant'' only changed after 8.0 and 10.0 minutes, respectively. The Spatial-model \emph{Soldier} (soldiers' distances to the rival Crystal) became more significant after 19.0 minutes compared to other Spatial-models, since soldiers became increasingly potent and important in the late-game stages in most MOBA esports, especially \emph{HoK}.

\subsection{Interpretability}
\label{sec:interpretability}

The proposed TSSTN model now serves as an essential module of an AI commentator system for \emph{HoK}.\footnote{A live-streaming video record of this system can be found at~\url{https://www.bilibili.com/video/BV1sK411W7JH}, and its English-subtitled record version can be found at the link~\url{https://drive.google.com/drive/folders/1oYT8DDlHOcA7_OXdyr4X3ErNDJXFN734?usp=sharing}. Our module works at game time 3:57, 5:04, and 8:25 in this record.} \textcolor{black}{Having been deployed for twelve months (since April 2020), this AI commentator can automatically generate real-time comments for \emph{HoK} games, and a large part of its comments rely on the interpretable prediction results of the proposed TSSTN model.} \textcolor{black}{Through the feedback of the audience and professional commentators, we believe that the TSSTN is able to provide human-interpretable information with satisfactory performance.} To illustrate the interpretability of the proposed TSSTN model, we exhibited a live-streaming game as an example, as Figure \ref{fig:example} shows. 

Specifically, the upper left part of this figure shows two real-time win probability curves for the two teams provided by the TSSTN model. The screenshots of three time-points in the live game are exhibited in the lower left, upper right and lower right parts, respectively. In the screenshots, the Chinese subtitles in white are real comments generated by the aforementioned \emph{HoK} AI commentator system, and the corresponding English translations are provided below. At time-point 1 (game time 0:03), since it was the beginning of the game, the TSSTN model predicted the Blue Team's win probability as $54.4\%$ based on the team composition (\emph{Heroes}) information only, so the AI commentator generated the comment: ``\textit{The game starts! The team composition of the Blue Team is in advantage; therefore, they get a win probability of 54.4\%}''.\footnote{\textcolor{black}{These comments are driven by rules written in the commentary system in advance and are triggered by the TSSTN's win predictions and interpretable results.}\label{comment}} Then, at time-point 2 (game time 2:11), due to the disadvantages in \emph{Gold} and \emph{Kill}, the Red Team's win probability dropped to $36.2\%$, and the contributions of feature groups ``Gold'' and ``Kill'' to the win prediction were -0.187 and -0.073, respectively. So the AI commentator generated the comment: ``\textit{The Red Team is at a disadvantage in terms of gold and kill-count. AI predicts that the Red Team’s win probability is only 36.2\%}''.\textsuperscript{\ref{comment}} Finally, at time-point 3 (game time 9:44), the Red Team reversed the win probability to $56.2\%$. This prediction was made based mainly on the difference of ``Gold'', ``Wild Resource'', and ``Tower'' between the two teams, so the AI generated the comment: ``\textit{With a real mess and bad start, the Red Team catches up gradually to kill more Tyrants and destroy more towers, thereby they accumulate an advantage in terms of Gold. Currently, their win probability has grown to 56.2\%}''.\textsuperscript{\ref{comment}}

Obviously, without the interpretable results, commentators were only able to make some trivial comments devoid of interest to the audience. For example, at time-point 3, one could only comment: ``\textit{Now the Red team reverses its situation and has a win probability of 56.2\%}'', which is less informative than the interpretable one made by the AI commentator with the help of the TSSTN model. Currently, the AI commentator generates 157.2 comments per game on average, and a large part of them depend on the proposed TSSTN model with a relatively even distribution over the six feature groups. This is a very significant characteristic that makes our AI commentator system different from other non-interpretable-prediction based systems.

\section{Conclusion}

In this paper, we proposed to predict the results of MOBA esports games in an interpretable way. A large-scale \emph{HoK} dataset containing abundant real-time MOBA game records was collected
to facilitate this study. Based on this dataset, we further proposed a Two-Stage Spatial-Temporal Network (TSSTN) to give interpretable intermediate results as well as reliable ultimate win predictions. The core idea of this model structure was to separate the effects of features' value differences and features' relative importance in order to decouple the contributions of different features. Our interpretable results can be utilized in various scenarios and promote the development of related industries. For example, the proposed TSSTN model has already been implemented in an AI commentator system for \emph{HoK} and achieved a sound performance. Our next goal is to construct automatic metrics to quantitatively evaluate the competence of different interpretable methods for MOBA esports win predictions. Further research can be done on improving the prediction accuracy and exploring the applications of interpretable prediction models in different industries.






%




\bibliographystyle{IEEEtran}
\bibliography{abc}
\end{document}